# Learning Efficient Structured Sparse Models

**Pablo Sprechmann**  sprec009@umn.edu
**Alex Bronstein**  bron@cs.technion.ac.il
Faculty of Engineering, Tel Aviv University, Ramat Aviv 69978, Israel

**Guillermo Sapiro**  guille@umn.edu
University of Minnesota - Department of Electrical and Computer Engineering, 200 Union Street SE, Minneapolis, USA

## Abstract

We present a comprehensive framework for structured sparse coding and modeling extending the recent ideas of using learnable fast regressors to approximate exact sparse codes. For this purpose, we propose an efficient feed forward architecture derived from the iteration of the block-coordinate algorithm. This architecture *approximates* the exact structured sparse codes with a fraction of the complexity of the standard optimization methods. We also show that by using different training objective functions, the proposed learnable sparse encoders are not only restricted to be approximants of the exact sparse code for a pre-given dictionary, but can be rather used as full-featured sparse encoders or even modelers. A simple implementation shows several orders of magnitude speedup compared to the state-of-the-art exact optimization algorithms at minimal performance degradation, making the proposed framework suitable for real time and large-scale applications.

## 1. Introduction

Sparse *coding* is the problem of representing signals as a sparse linear combination of elementary atoms of a given dictionary. Sparse *modeling* aims at learning such (non-)parametric dictionaries from the data themselves. In addition to being very attractive at the theoretical level, a large class of signals is well described by this model, as demonstrated by numerous state-of-the-art results in diverse applications.

The main challenge of all optimization-based sparse coding and modeling approaches is their relatively high computational complexity. Consequently, a significant amount of effort has been devoted to developing efficient optimization schemes (Beck & Teboulle, 2009; Li & Osher, 2009; Nesterov, 2007; Xiang et al., 2011). Despite the permanent progress reported in the literature, the state-of-the-art algorithms require tens or hundreds of iterations to converge, making them infeasible for real-time or very large (modern size) applications.

Recent works have proposed to trade off precision in the sparse representation for computational speedup (Jarrett et al., 2009; Kavukcuoglu et al., 2010), by learning non-linear regressors capable of producing good approximations of sparse codes in a fixed amount of time. The insightful work by (Gregor & LeCun, 2010) introduced a novel approach in which the regressors are multi-layer artificial neural networks (NN) with a particular architecture inspired by successful optimization algorithms for solving sparse coding problems. These regressors are trained to minimize the MSE between the predicted and exact codes over a given training set. Unlike previous predictive approaches, the system introduced in (Gregor & LeCun, 2010) has an architecture capable of producing more accurate approximations of the true sparse codes, since it allows an approximate "explaining away" to take place during inference (see (Gregor & LeCun, 2010) for details). In this paper we propose several extensions of (Gregor & LeCun, 2010), including the consideration of more general sparse coding paradigms (hierarchical and non-overlapping grouped), adding online adaptation of the underlying dictionary/model, thereby extending the applicability of this fast encoding framework. The proposed approach can be used with a predefined dictionary or learn it in an online manner on the very same data vectors fed to it.

While differently motivated, in the case in which the dictionary is learned, the framework is related to recent efforts in producing NN based sparse representations, see (Goodfellow et al., 2009; Ranzato et al., 2007) and references therein. It can be interpreted as an online trainable sparse auto-encoder (Goodfellow et al., 2009) with a sophisticated encoder and simple linear decoder. The higher complexity of the proposed architecture in the encoder allows the system to produce accurate estimates of true structured sparse codes.

In Section 2 we briefly present the general problem of hierarchical structured sparse coding and in Section 3 discuss the optimization algorithm used to inspire the architecture of the encoders. In Section 4 we present the new sparse encoders and the new objective functions used for their training. Experimental results in real audio and image analysis tasks are presented in Section 5. Finally, conclusions are drawn in Section 6.

## 2. Structured Sparse Models

The underlying assumption of sparse models is that the input vectors can be reconstructed accurately as a linear combination of some (usually learned) basis vectors (factors or dictionary atoms) with a small number of non-zero coefficients. *Structured* sparse models further assume that the pattern of non-zero coefficients exhibits a specific structure known *a priori*.

Let $\mathbf{D} \in \mathbb{R}^{m \times p}$ be a dictionary with $p$ $m$-dimensional atoms. We define groups of atoms through their indexes, $G \subseteq \{1, \ldots, p\}$. Then, we define a group structure, $\mathcal{G}$, as collection of groups of atoms, $\mathcal{G} = \{G_1, \ldots, G_{|\mathcal{G}|}\}$. For an input vector $\mathbf{x} \in \mathbb{R}^m$, the corresponding structured sparse code, $\mathbf{z} \in \mathbb{R}^p$, associated to the group structure $\mathcal{G}$, can be obtained by solving the convex program,

$$\min_{\mathbf{z} \in \mathbb{R}^p} \frac{1}{2} \|\mathbf{x} - \mathbf{D}\mathbf{z}\|_2^2 + \psi(\mathbf{z}), \quad (1)$$

$$\psi(\mathbf{z}) = \sum_{r \in \mathcal{G}} \lambda_r \|\mathbf{z}_r\|_2, \quad (2)$$

where the vector $\mathbf{z}_r \in \mathbb{R}^{|G_r|}$ contains the coefficients of $\mathbf{z}$ belonging to group $r$, and $\lambda_r$ are scalar weights controlling the sparsity level.

The regularizer function $\psi$ in (1) can be seen as a generalization of the $\ell_1$ regularizer used in standard sparse coding, as the latter arises from the special case of singleton groups $\mathcal{G} = \{\{1\}, \{2\}, \ldots, \{p\}\}$ and setting $\lambda_r = 1$. As such, its effect on the groups of $\mathbf{z}$ is a natural generalization of the one obtained with standard sparse coding: it "turns on" and "off" atoms in groups according to the structure imposed by $\mathcal{G}$.

**Algorithm 1** Forward-backward splitting method.
**input** : Data $\mathbf{x}$, dictionary $\mathbf{D}$, regularizer $\psi$.
**output**: Sparse code $\mathbf{z}$.
Define $\mathbf{S} = \mathbf{I} - \frac{1}{\alpha}\mathbf{D}^{\mathrm{T}}\mathbf{D}$, $\mathbf{W} = \frac{1}{\alpha}\mathbf{D}^{\mathrm{T}}$, $t = \frac{1}{\alpha}$.
Initialize $\mathbf{z} = \mathbf{0}$ and $\mathbf{b} = \mathbf{W}\mathbf{x}$.
**repeat**
$\quad \mathbf{z} = \mathrm{prox}_{t\psi}(\mathbf{b})$
$\quad \mathbf{b} = \mathbf{b} + \mathbf{S}\mathbf{z}$
**until** *until convergence*;

Several important structured sparsity settings can be cast as particular cases of (1): *sparse coding*, as mentioned above, which is often referred to as Lasso (Tibshirani, 1996) or basis pursuit (Chen et al., 1999; Donoho, 2006); *group sparse coding*, a generalization of the standard sparse coding to the cases in which the dictionary is sub-divided into groups that are known to be active or inactive simultaneously (Yuan & Lin, 2006), in this case $\mathcal{G}$ is a partition of $\{1, \ldots, p\}$; *hierarchical sparse coding*, assuming a hierarchical structure of the non-zero coefficients (Zhao et al., 2009; Jenatton et al., 2011; Sprechmann et al., 2011). The groups in $\mathcal{G}$ form a hierarchy with respect to the inclusion relation (a tree structure), that is, if two groups overlap, then one is completely included in the other one; and *collaborative sparse coding* generalizing the concept of structured sparse coding to collections of input vectors by promoting given patterns of non-zero elements in the coefficient matrix (Eldar & Rauhut, 2010; Sprechmann et al., 2011).

## 3. Optimization Algorithms

State-of-the-art approaches for solving (1) rely on the family of proximal splitting methods (see (Bach et al., 2011) and references therein). Next, we briefly introduce proximal methods and an algorithm for solving hierarchical sparse coding problems (Tseng, 2001) that will be used to construct trainable sparse encoders.

### 3.1. Forward-Backward Splitting

The forward-backward splitting method is designed for solving unconstrained optimization problems in which the cost function can be split as

$$\min_{\mathbf{z} \in \mathbb{R}^m} f_1(\mathbf{z}) + f_2(\mathbf{z}), \quad (3)$$

where $f_1$ is convex and differentiable with a $\frac{1}{\alpha}$-Lipschitz continuous gradient, and $f_2$ is convex extended real valued and possibly non-smooth. Clearly, problem (1) falls in this category by considering $f_1(\mathbf{z}) = \frac{1}{2}\|\mathbf{x} - \mathbf{D}\mathbf{z}\|_2^2$ and $f_2(\mathbf{z}) = \psi(\mathbf{z})$.

The forward-backward splitting method with fixed constant step defines a series of iterates,

$$\mathbf{z}^{k+1} = \text{prox}_{\alpha f_2}(\mathbf{z}^k - \frac{1}{\alpha}\nabla f_1(\mathbf{z}^k)), \quad (4)$$

where $\text{prox}_{f_2}(\mathbf{z}) = \underset{\mathbf{u} \in \mathbb{R}^m}{\text{argmin}} ||\mathbf{u} - \mathbf{z}||_2^2 + f_2(\mathbf{u})$ denotes the proximal operator of $f_2$. The procedure is given in Algorithm 1.

The forward-backward method becomes particularly interesting when the proximal operator of $\psi$ can be computed exactly and efficiently, e.g., in standard or group-structured sparse coding. When the groups of $\mathcal{G}$ overlap arbitrarily, there is no efficient way of doing so directly. However, there exist important exceptions such as the hierarchical setting with tree-structured groups which is discussed in the sequel. Accelerated versions of proximal methods have been largely studied in the literature to improve their convergence rate (Beck & Teboulle, 2009; Nesterov, 2007). While these variants are the fastest exact solvers available (both in theory and practice), they still require tens or hundreds of iterations to achieve convergence. In the following sections we will elaborate in the standard versions of the algorithm since we are only interested in constructing an architecture for the proposed sparse encoders.

### 3.2. Proximal Operators

To simplify the notation, we will henceforth formulate all the derivations for the case of two-level hierarchical sparse coding, referred as HiLasso (Friedman et al., 2010; Sprechmann et al., 2011). This captures the essence of hierarchical sparse models and the generalization to more layers (Jenatton et al., 2011) or to a collaborative scheme (Sprechmann et al., 2011) is straightforward.

The HiLasso model was introduced for simultaneously promoting sparsity at both, group and coefficient level. Given a partition $\mathcal{P} = \{G_1, \ldots, G_{|\mathcal{P}|}\}$, the group structure $\mathcal{G}$ can be expressed as the union of two partitions: $\mathcal{P}$ and the set of singletons. Thus, the regularizer $\psi$ becomes

$$\psi(\mathbf{z}) = \sum_{j=1}^{p} \lambda_j \|\mathbf{z}_j\|_1 + \sum_{r=1}^{|\mathcal{P}|} \mu_r \|\mathbf{z}_r\|_2. \quad (5)$$

The proximal operator of $\psi$ can then be computed in closed form. Given a partition of the group of indexes, $\mathcal{P}$, and a vector of thresholding parameters $\boldsymbol{\lambda} \in \mathbb{R}^{|\mathcal{P}|}$, we define the group separable operator $\boldsymbol{\pi}_{\boldsymbol{\lambda}} : \mathbb{R}^p \to \mathbb{R}^p$

**Algorithm 2** BCoFB algorithm.
**input** : Data $\mathbf{x}$, structured dictionary $\mathbf{D}$, $\boldsymbol{\lambda}$, $\boldsymbol{\mu}$.
**output**: Structured sparse code $\mathbf{z}$.
Bound Lipschitz constant $\alpha \leq \max_r \|\mathbf{D}_r\|_2^2$
Define $\mathbf{S} = \mathbf{I} - \frac{1}{\alpha}\mathbf{D}^T\mathbf{D}$, $\mathbf{W} = \frac{1}{\alpha}\mathbf{D}^T$, $\mathbf{s} = \frac{1}{\alpha}\boldsymbol{\mu}$, and $\mathbf{t} = \frac{1}{\alpha}\boldsymbol{\lambda}$.
Initialize $\mathbf{z} = \mathbf{0}$ and $\mathbf{b} = \mathbf{W}\mathbf{x}$.
**repeat**
$\quad \mathbf{y} = \boldsymbol{\pi}_{\mathbf{s},\mathbf{t}}(\mathbf{b})$
$\quad \mathbf{e} = \mathbf{y} - \mathbf{z}$
$\quad g = \arg\max_r \|\mathbf{e}_r\|_2$
$\quad \mathbf{b} = \mathbf{b} + \mathbf{S}_g \mathbf{e}_g$
$\quad \mathbf{z}_g = \mathbf{y}_g$
**until** *until convergence*;
Output $\mathbf{z} = \boldsymbol{\pi}_{\mathbf{s},\mathbf{t}}(\mathbf{b})$

for $r = 1, \ldots, |\mathcal{P}|$ as

$$\boldsymbol{\pi}_{\boldsymbol{\lambda}}(\mathbf{z})_r = \frac{\max\{0, \|\mathbf{z}_r\|_2 - \lambda_r\}}{\|\mathbf{z}_r\|_2}\mathbf{z}_r \quad (6)$$

if $\|\mathbf{z}_r\|_2 > 0$, and $\mathbf{0}$ otherwise. Note that $\boldsymbol{\pi}_{\boldsymbol{\lambda}}$ applies a vector soft-thresholding to each group in $\mathcal{P}$. The proximal operator of (5) can be expressed as (Jenatton et al., 2011; Sprechmann et al., 2011),

$$\boldsymbol{\pi}_{\boldsymbol{\lambda},\boldsymbol{\mu}}(\mathbf{z}) = \boldsymbol{\pi}_{\boldsymbol{\mu}}(\boldsymbol{\pi}_{\boldsymbol{\lambda}}(\mathbf{z})), \quad (7)$$

a composition of the proximal operators associated to each of the partitions in $\mathcal{G}$: $\mathcal{P}$ and the set of singletons.

The Lasso problem is a particular case of HiLasso with $\boldsymbol{\mu} = \mathbf{0}$ and $\boldsymbol{\lambda} = \lambda \mathbf{1}$, in which the proximal operator in (7) reduces to the scalar soft-thresholding operator and Algorithm 1 corresponds then to the popular iterative shrinkage and thresholding algorithm (ISTA) (Beck & Teboulle, 2009).

### 3.3. Block-Coordinate Forward-Backward Algorithm

In Algorithm 1, every iteration requires the update of all the groups of coefficients in the partition $\mathcal{P}$, according to (7). One can choose a block coordinate strategy where only one block is updated at a time (Tseng, 2001). In this paper we will refer to this algorithm as Block-Coordinate Forward-Backward algorithm (BCoFB) (see (Bach et al., 2011) for a review on similar algorithms). The iterates of BCoFB are,

$$\begin{aligned}\mathbf{p} &= \boldsymbol{\pi}_{\boldsymbol{\lambda},\boldsymbol{\mu}}(\mathbf{z}^k), \\ \mathbf{z}^{k+1} &= \mathbf{z}^k, \\ \mathbf{z}_g^{k+1} &= \mathbf{p}_g - \frac{1}{\alpha}\mathbf{D}_g^T(\mathbf{D}_g\mathbf{p}_g^k - \mathbf{x}_g),\end{aligned} \quad (8)$$

where again here $1/\alpha$ is the Lipschitz constant of the fitting term and $g$ is the index of the group in $\mathcal{P}$ to be updated at the $k$-th iteration, according to some selection rule. Inspired by the coordinate descent algorithm (CoD) introduced for standard sparse coding in (Li & Osher, 2009), we propose an heuristic variant of BCoFB algorithm, that updates the group

$$g = \underset{j}{\operatorname{argmax}} \|\mathbf{z}_j^{k+1} - \mathbf{z}_j^k\|_2^2.$$

It can be shown that this quantity provides a lower bound in the decrease of the cost function for each possible group update. The procedure is summarized in Algorithm 2.

In the case of standard sparsity, Algorithm 2 with $\alpha = 1$ is identical to CoD (Li & Osher, 2009). This algorithm was used in (Gregor & LeCun, 2010) to build trainable sparse encoders.

## 4. Fast Structured Sparse Encoders

In order to make sparse coding feasible in real time settings, it has been recently proposed to learn non-linear regressors capable of producing good approximations of sparse codes in a fixed amount of time (Jarrett et al., 2009; Kavukcuoglu et al., 2010). The main idea is to construct a parametric regressor $\mathbf{h}(\mathbf{x}, \boldsymbol{\Theta})$, with some set of parameters, collectively denoted as $\boldsymbol{\Theta}$, that minimizes the loss function

$$\mathcal{L}(\boldsymbol{\Theta}) = \frac{1}{N} \sum_n L(\boldsymbol{\Theta}, \mathbf{x}_n) \tag{9}$$

on a training set $\{\mathbf{x}_1, \ldots, \mathbf{x}_N\}$. Here, $L(\boldsymbol{\Theta}, \mathbf{x}_n) = \frac{1}{2}\|\mathbf{z}_n^* - \mathbf{z}_n\|_2^2$, $\mathbf{z}_n^*$ is the exact sparse code of $\mathbf{x}_n$ obtained by solving the Lasso problem, and $\mathbf{z}_n = \mathbf{h}(\mathbf{x}_n, \boldsymbol{\Theta})$ is its approximation. While this setting is very generic, the application of off-the-shelf regressors has been later shown to produce relatively low-quality approximations (Gregor & LeCun, 2010).

(Gregor & LeCun, 2010) proposed then two particular regressors implemented as a truncated form of ISTA and CoD algorithms. Essentially, these regressors are multi-layer artificial NN's where each layer implements a single iteration of ISTA or CoD. For example, in the CoD architecture, the learned parameters of the network are the matrices $\mathbf{S}$ and $\mathbf{W}$, and the set of element-wise thresholds $\mathbf{t}$.

Naturally, as an alternative to learning, one could simply set the parameters $\mathbf{S}$, $\mathbf{W}$, and $\mathbf{t}$ as prescribed by the CoD algorithm (a particular case of Algorithm 2), terminating it after a small number of iterations. However, it is by no means guaranteed that such a truncated CoD algorithm will produce the best sparse code approximation with the same (small) number of layers; in practice, without the learning, such truncated approximations are typically useless. Still, even when learning the parameters, it is hopeless to expect the NN regressor to produce good sparse codes for *any* input data. Yet, (Gregor & LeCun, 2010) showed that the network does approximate well sparse codes for input vectors coming from the same distribution as the one used in training.

Another remarkable property of the ISTA and CoD sparse encoder architectures is that they are continuous and almost everywhere $\mathcal{C}^1$ with respect to the parameters and the inputs. Differentiability with respect to the parameters allows the use of (sub)gradient descent methods for training, while differentiability with respect to the inputs allows backpropagation of the gradients and the use of the sparse encoders as modules in bigger globally-trained systems.

The minimization of a loss function $\mathcal{L}(\boldsymbol{\Theta})$ with respect to $\boldsymbol{\Theta}$ requires the computation of the (sub)gradients $d\mathcal{L}(\boldsymbol{\Theta}, \mathbf{x}_n)/d\boldsymbol{\Theta}$, which is achieved by the backpropagation procedure. Backpropagation starts with differentiating $\mathcal{L}(\boldsymbol{\Theta}, \mathbf{x}_n)$ with respect to the output of the last network layer, and propagating the (sub)gradients down to the input layer, multiplying them by the Jacobian matrices of the traversed layers.

### 4.1. Hierarchical Sparse Encoders

We now extend Gregor&LeCun's idea to hierarchical (structured) sparse code regressors. We consider a feed-forward architecture based on the BCoFB, where each layer implements a single iteration of the BCoFB proximal method (Algorithm 2). The encoder architecture is depicted in Figure 1. Each layer essentially consists of the nonlinear proximal operator $\boldsymbol{\pi}_{\mathbf{s},\mathbf{t}}$ followed by a group selector and a linear operation $\mathbf{S}_g$ corresponding to that group. The network parameters are initialized as in Algorithm 2. In the particular case of $\alpha = 1$ and $\mathbf{s} = \mathbf{0}$, the CoD architecture is obtained.

### 4.2. Alternative Training Objective Functions

So far, we have followed Gregor&LeCun in considering NN encoders as regressors whose only role is to reproduce as faithfully as possible the ideal sparse codes $\mathbf{z}_n^*$ produced by an iterative sparse coding algorithm (e.g., Lasso or HiLasso). This is achieved by training the networks to minimize the $\ell_2$ discrepancy between the outputs of the network and the corresponding $\mathbf{z}_n^*$. We propose to consider the neural network sparse coders (both structured and unstructured) not as regressors approximating an iterative algorithm, but as full-featured sparse encoders (even modelers) in their own right. To achieve this paradigm shift, we abandon the ideal sparse codes and introduce alternative training objectives as detailed in the sequel.

A general sparse coding problem can be viewed as a mapping between a data vector $\mathbf{x}$ and the corresponding sparse code $\mathbf{z}$ minimizing an aggregate of a fitting term and a (possibly, structured) regularizer, $f(\mathbf{x}, \mathbf{z}) = \frac{1}{2} \|\mathbf{x} - \mathbf{D}\mathbf{z}\|_2^2 + \psi(\mathbf{z})$. Since the latter objective is trusted as an indication of the code quality, we can train the network to minimize the ensemble average of $f$ on a training set with $\mathbf{z} = \arg\min f(\mathbf{x}, \mathbf{z})$ replaced by $\mathbf{z} = \mathbf{h}(\mathbf{x}, \mathbf{\Theta})$, obtaining the objective

$$\mathcal{L}(\mathbf{\Theta}) = \frac{1}{N} \sum_n f(\mathbf{x}_n, \mathbf{h}(\mathbf{x}_n, \mathbf{\Theta})). \quad (10)$$

Given an application, one therefore has to select an objective with an appropriate regularizer $\psi$ corresponding to the problem structure, and a sparse encoder architecture consistent with that structure, and train the latter to minimize the objective on a representative set of data vectors. We found that selecting the sparse encoder with the structure consistent with the training objective and the inherent structure of the problem is crucial for the production of high-quality sparse codes.

While sparse encoders based on NN's are trained by minimizing a non-convex function on a training set, and are therefore prone to local convergence and overfitting, we can argue that in most practical problems, the dictionary $\mathbf{D}$ is also found by solving a non-convex dictionary learning problem based on a representative data distribution. Consequently, unless the dictionary is constructed using some domain knowledge, the use of NN sparse encoders is not conceptually different from using iterative sparse modeling algorithms.

Furthermore, one can consider the dictionary as another optimization variable in the training, and minimize $\mathcal{L}$ with respect to both $\mathbf{D}$ and the network parameters $\mathbf{\Theta}$, alternating between network training and dictionary update iterations. This essentially extends the proposed efficient sparse coding framework into full-featured sparse modeling, as detailed next.

### 4.3. Online Learning

Interpreting the NN's as standalone sparse encoders and removing the reference exact sparse codes makes the training problem completely unsupervised. Consequently, one may train the network (and possibly adapt the dictionary as well) on the very same data vectors fed to it for sparse coding. This allows using the proposed framework in online learning applications. A full online sparse modeling scenario consists of (a) initializing the dictionary (e.g., by a random sample of the initially observed training data vectors); (b) fixing the dictionary in the training objective and adapting the network parameters to the newly arriving data using an online learning algorithm (we use an online version of stochastic gradient in small batches as detailed in Section 5); and (c) fixing the sparse codes and adapting the dictionary using an online dictionary learning algorithm (e.g., (Mairal et al., 2009)). Note that all the above stages are completely free of iterative sparse coding, which translates into low latency computational complexity allowing real time applications.

### 4.4. Supervised and Discriminative Learning

The proposed sparse modeling framework allows to naturally incorporate side information about training data vectors, making the learning supervised. Space limitations prevent us from elaborating on this setting; in what follows, we outline several examples leaving the details to the extended version of this paper.

In the group or hierarchical Lasso case, one may know for each data vector the desired active groups. Incorporating this information into the training objective is possible by using $\psi$ as in (5) with $\mu_r$ set separately for each training vector $\mathbf{x}_n$ to low values to promote the activation of a knowingly active group $r$, or to high values to discourage the activation of a knowingly inactive group.

In other applications, data vectors can come in pairs of knowingly similar or dissimilar vectors, and one may want to minimize some natural distance between sparse codes of the similar vectors, while maximizing the distance on the dissimilar ones. This scenario is of particular interest in retrieval applications, where sparse data representations are desirable due to their amenability to efficient indexing. Incorporating such a *similarity preservation* term into the training objective is common practice in metric learning (see, e.g., (Weinberger & Saul, 2009)), but is challenging in sparse coding due to the fact that when the sparse codes are produced by an iterative algorithm, one faces the problem of minimizing a training objective $\mathcal{L}$ depending on the minimizers of another objective $f$. When using the NN sparse modelers instead, the training is handled using standard methods.

Finally, in many applications the data do not have Euclidean structure and supervised learning can be used to construct an optimal discriminative metric. This can be achieved, for example, by replacing the Euclidean fitting term with the Mahalanobis counterpart, $\|\mathbf{x} - \mathbf{D}\mathbf{x}\|_{\mathbf{Q}^\mathsf{T}\mathbf{Q}}^2 = \|\mathbf{Q}(\mathbf{x} - \mathbf{D}\mathbf{x})\|_2^2$, where $\mathbf{Q}$ is a discriminative projection matrix. In such scenarios, it is desirable to combine sparse modeling with metric learning. This problem has not been considered before as it is impractical when the sparse codes are obtained by minimization of $f$. It does become practical, however, when NN encoders are used instead.

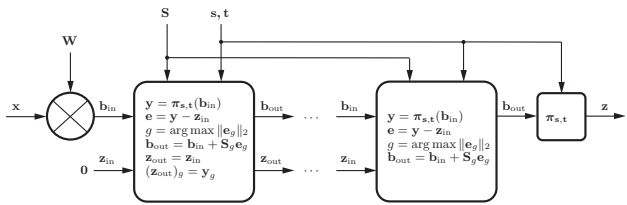

*Figure 1.* BCoFB structured sparse encoder architecture with two levels of hierarchy (a "HiLasso" network).

*Table 1.* Misclassification rates on MNIST digits.

| Code | Dictionary size | |
|---|---|---|
| | 100 | 289 |
| NN G-L | 3.76% | 5.98% |
| NN Lasso | 2.65% | 2.51% |
| Exact Lasso | 1.99% | 1.47% |

## 5. Experimental Results

All NN's were implemented in Matlab with built-in GPU acceleration and executed on state-of-the-art Intel Xeon E5620 CPU and NVIDIA Tesla C2070 GPU. Even with this by no means optimized code, the propagation of $10^5$ 100-dimensional vector through a 10-layer structured network with the proposed BCoFB architecture takes only 3.6 seconds, which is equivalent to $3.6\mu sec$ spent per vector per layer. This is several orders of magnitude faster than the exceptionally optimized multithreaded SPAMS HiLasso code (Jenatton et al., 2011) executed on the CPU. Such benefits of parallelization are possible due to the fixed datapath and complexity of the NN encoder compared to the iterative solver.

In all experiments, training was performed using gradient descent safeguarded by Armijo rule. We refer as *NN G-L* to the NN sparse encoders obtained by minimizing Gregor&LeCun's objective function, this is, the $\ell_2$ error with the output of the exact encoder. It will be explicitly stated when NN sparse encoders are trained using a specific objective function (e.g., *NN Lasso*).

### 5.1. Classification

In this experiment, we evaluate the performance of unstructured NN sparse encoders in the MNIST digit classification task. The MNIST images were resampled to $17 \times 17$ (289-dimensional) patches. A set of ten dictionaries was trained for each class. Classification was performed by encoding a test vector in each of the dictionaries and assigning the label corresponding to the smallest value of the full Lasso objective.

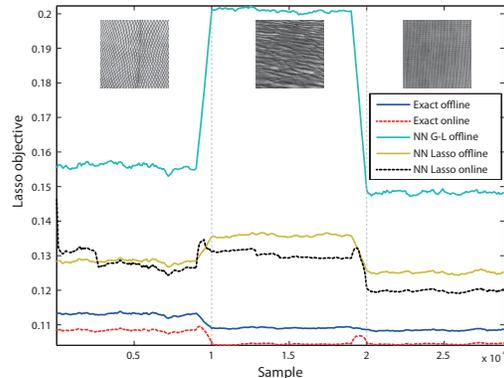

*Figure 2.* Performance of different sparse encoders measured using the Lasso objective as the function of sample number in the online learning experiment. Shown are the three groups of patches corresponding to different texture images from the Brodatz dataset.

The following sparse coders were compared: exact sparse codes (*Exact Lasso*), unstructured *NN G-L*, and unstructured *NN Lasso* (a CoD network trained using the Lasso objective). Ten networks were trained, one per each class; all contained $T = 5$ CoD layers. $\lambda = 0.1$ was used in the Lasso objective. Dictionaries with 100 (under-complete) and 289 (complete) atoms were used. Further increase in the dictionary size did not exhibit significant performance improvement.

Table 1 summarizes the misclassification rates of each of the sparse encoders. Performance of the *NN G-L* sparse encoder decreases with the increase of the dictionary size, while the discrepancy with the exact codes drops. On the other hand, better performance in terms of the Lasso objective consistently correlates with better classification rates, which makes *NN Lasso* a more favorable choice. Dictionary adaptation in the training of the *NN Lasso* encoder brings only a small improvement in performance, diminishing with the dictionary size. We attribute this to the relative low complexity of the data.

### 5.2. Online Learning

In this experiment, we evaluate the online learning capabilities of unstructured NN sparse encoders. As the input data we used $30 \times 10^4$ randomly located $8 \times 8$ patches from three images from the Brodatz texture dataset (Randen & Husoy, 1999). The patches were ordered in three consecutive blocks of $10^4$ patches from each image. Dictionary size was fixed to 64 atoms. $\lambda = 1$ was used in the Lasso objective.

Online learning was performed in overlapping windows of $1,000$ vectors with a step of 100 vectors. We com-pared standard online dictionary learning (*Exact Lasso online*) with unstructured *NN Lasso* with dictionary adaptation in a given window (*NN Lasso online*), initialized by the network parameters from the previous windows. In the latter case, the dictionary was initialized by a random subset of 64 out of the first 1,000 data vectors (therefore, no iterative sparse coding). As the reference, we also compared the following three offline algorithms trained on a distinct set of 6,000 patches extracted from the same images: standard dictionary learning (*Exact Lasso offline*); unstructured *NN G-L* (*NN G-L offline*), and unstructured *NN Lasso* (*NN Lasso offline*). All NN's used $T = 4$ CoD layers.

Table 2. Misclassification rates on the audio dataset.

| Code | Error rate |
|---|---|
| NN G-L unstructured | 6.08% |
| NN G-L structured | 3.53% |
| NN discriminative structured | 3.44% |
| Exact | 2.35% |

Performance measured in terms of the Lasso objective is reported in Figure 2. Exact offline sparse encoder achieved the best results among all offline encoders. It is, however, outperformed by the exact online encoder due to its ability to adapt the dictionary to a specific class of data. The performance of the *NN Lasso online* encoder is slightly inferior to the *Exact Lasso offline*; the online version performs better after the network parameters and the dictionary adapt to the current class of data. Finally, the *NN G-L offline* encoder has the lowest, significantly inferior performance.

This experiment shows that, while the drop in performance compared to the exact encoder is relatively low, the computational complexity of the *NN Lasso online* encoder is tremendously lower and fixed.

### 5.3. Structured Coding

We first evaluate the performance of the structured sparse encoders in a speaker identification task reproduced from (Sprechmann et al., 2011). In this application the authors use HiLasso to automatically detect the present speakers in a given mixed signal. We repeat this experiments using the proposed efficient structured sparse encoders instead.

The dataset consists of recordings of five different radio speakers, two females and three males. 25% of the samples was used for training, and the rest for testing. Within the testing data, two sets of waveforms were created: one containing isolated speakers, and another containing all possible combinations of mixtures of two speakers. Signals are decomposed into a set of overlapping local time frames of 512 samples with 75% overlap, such that the properties of the signal remain stable within each frame. An 80-dimensional feature vector is obtained for each audio frame as its short-time power spectrum envelope (refer to (Sprechmann et al., 2011) for details). Five undercomplete dictionaries with 50 atoms were trained on the single speaker set minimizing the Lasso objective with $\lambda = 0.2$ (one dictionary per speaker), and then combined into a single structured dictionary containing 250 atoms. Increasing the dictionary size exhibited negligible performance benefits. Speaker identification was performed by first encoding a test vector in the structured dictionary and measuring the $\ell_2$ energy of each of the five groups. Energies were sum-pooled over 500 time samples selecting the labels of the highest two.

The following structured sparse encoders were compared: exact HiLasso codes with $\mu = 0.05$ (*Exact*), unstructured *NN G-L* trained on the exact HiLasso codes (*NN G-L unstructured*), structured *NN G-L* trained on the same codes (*NN G-L structured*), and a structured network with a discriminative cost function with regularization term in which the weights $\mu_r$ were set independently for each data vector to $-1$ or $1$ to promote or discourage the activation of groups corresponding to knowingly active or silent speakers respectively, (*NN discriminative structured*). All NN's used the same single structured dictionary and contained $T = 2$ layers with the BCoFB architecture.

Table 2 summarizes the obtained misclassification rates. It is remarkable that using a structured architecture instead of its unstructured counterpart with the same number of layers and the same dictionary increases performance by nearly a factor of two. The use of the discriminative objective further improves performance. Surprisingly, using NN's with only two layers cedes just about 1% of correct classification rate.

The structured architecture showed a crucial roll in producing accurate structured sparse codes. We now show that this observation is also valid in a more general setting. We repeated the same experiment as before but with randomly generated synthetic data that truly has a structure sparse representation under a given dictionary (unknown for the NN's). Results are summarized in Figure 3.

## 6. Conclusion

Marrying ideas from convex optimization with multi-layer neural networks, we have developed in this work a comprehensive framework for modern sparse mod-eling for real time and large scale applications. The framework includes different objective functions, from reconstruction to classification, allows different sparse coding structures from hierarchical to group similarity, and addresses online learning scenarios. A simple implementation already achieves several order of magnitude speedups when compared to the state-of-the-art, at minimal cost in performance, opening the door for practical algorithms following the demonstrated success of sparse modeling in various applications.

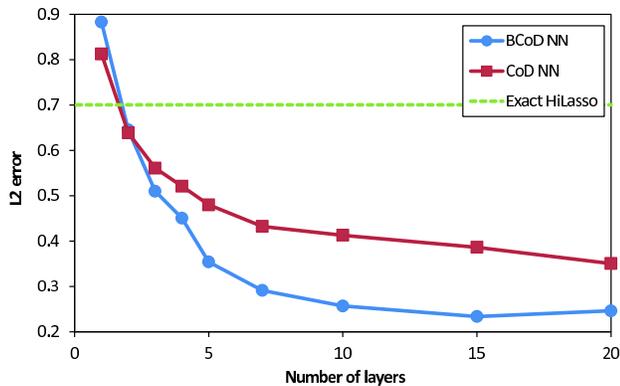

Figure 3. Performance ($\ell_2$ error) of structured and unstructured NN on structured sparse synthetic data.

An extension of the proposed approach to other structured sparse modeling problems such as robust PCA and non-negative matrix factorization is available at http://www.eng.tau.ac.il/~bron/publications_conference.html and will be published elsewhere due to lack of space.

## Acknowledgement

This research was supported in part by ONR, NGA, ARO, NSF, NSSEFF, and BSF.

## References

Bach, F., Jenatton, R., Mairal, J., and Obozinski, G. Convex optimization with sparsity-inducing norms. In *Optimization for Machine Learning*. MIT Press, 2011.

Beck, A. and Teboulle, M. A fast iterative shrinkage-thresholding algorithm for linear inverse problems. *SIAM J. Img. Sci.*, 2:183–202, March 2009.

Chen, S., Donoho, D., and Saunders, M. Atomic decomposition by basis pursuit. *SIAM J. Scientific Computing*, 20(1):33–61, 1999.

Donoho, D. Compressed sensing. *IEEE Trans. on Inf. Theory*, 52(4):1289–1306, Apr. 2006.

Eldar, Y. C. and Rauhut, H. Average case analysis of multichannel sparse recovery using convex relaxation. *IEEE Trans. on Inf. Theory*, 56(1):505–519, 2010.

Friedman, J., Hastie, T., and Tibshirani, R. A note on the group lasso and a sparse group lasso. Preprint, 2010.

Goodfellow, I., Le, Q., Saxe, A., Lee, H., and Ng, A. Y. Measuring invariances in deep networks. In *In NIPS*, pp. 646–654. 2009.

Gregor, K. and LeCun, Y. Learning fast approximations of sparse coding. In *ICML*, pp. 399–406, 2010.

Jarrett, K., Kavukcuoglu, K., Ranzato, M.A., and LeCun, Y. What is the best multi-stage architecture for object recognition? In *CVPR*, pp. 2146–2153, 2009.

Jenatton, R., Mairal, J., Obozinski, G., and Bach, F. Proximal methods for hierarchical sparse coding. *JMLR*, 12:2297–2334, 2011.

Kavukcuoglu, K., Ranzato, M.A., and LeCun, Y. Fast inference in sparse coding algorithms with applications to object recognition. *arXiv:1010.3467*, 2010.

Li, Y. and Osher, S. Coordinate descent optimization for $\ell_1$ minimization with application to compressed sensing; a greedy algorithm. *Inverse Problems and Imaging*, 3:487–503, 2009.

Mairal, J., Bach, F., Ponce, J., and Sapiro, G. Online dictionary learning for sparse coding. In *ICML*, pp. 689–696, 2009.

Nesterov, Y. Gradient methods for minimizing composite objective function. In *CORE*. Catholic University of Louvain, Louvain-la-Neuve, Belgium, 2007.

Randen, T. and Husoy, J. H. Filtering for texture classification: a comparative study. *IEEE Trans. Pattern Anal. Mach. Intell.*, 21(4):291–310, 1999.

Ranzato, M., Huang, F. J., Boureau, Y-L., and LeCun, Y. Unsupervised learning of invariant feature hierarchies with applications to object recognition. In *CVPR*, 2007.

Sprechmann, P., Ramírez, I., Sapiro, G., and Eldar, Y. C. C-hilasso: A collaborative hierarchical sparse modeling framework. *IEEE Trans. Signal Process.*, 59(9):4183–4198, 2011.

Tibshirani, R. Regression shrinkage and selection via the LASSO. *J. Royal Stat. Society: Series B*, 58(1):267–288, 1996.

Tseng, P. Convergence of a block coordinate descent method for nondifferentiable minimization. *J. Optim. Theory Appl.*, 109(3):475–494, June 2001.

Weinberger, K.Q. and Saul, L.K. Distance metric learning for large margin nearest neighbor classification. *JMLR*, 10:207–244, 2009.

Xiang, Z. J., Xu, H., and Ramadge, P. J. Learning sparse representations of high dimensional data on large scale dictionaries. *In NIPS*, 24:900–908, 2011.

Yuan, M. and Lin, Y. Model selection and estimation in regression with grouped variables. *J. Royal Stat. Society, Series B*, 68:49–67, 2006.

Zhao, P., Rocha, G., and Yu, B. The composite absolute penalties family for grouped and hierarchical variable selection. *Annals of Statistics*, 37(6A):3468, 2009.